\documentclass{IEEEtran}  
\usepackage[utf8]{inputenc}
\usepackage{bbm}
\usepackage{amsmath}
\usepackage{geometry}
\usepackage{graphicx}
\usepackage{booktabs}
\usepackage{svg}
\usepackage{float}
\title{Improving Learning from Demonstrations by Learning from Experience}
\author{Liu Haofeng, Chen Yiwen, Tan Jiayi, Marcelo H Ang}

\date{August 2021}

\begin{document}
\maketitle

\begin{abstract}
\normalsize How to make imitation learning more general with relatively limited demonstrations has been a persistent problem in reinforcement learning. Poor demonstrations lead to narrow and biased data distribution. The non-Markovian human expert demonstration presents agents with a learning challenge, and over-reliance on sub-optimal trajectories makes it hard to improve its performance. To solve these problems, we propose a new algorithm named TD3fG (TD3 learning from a generator) that can smoothly transition from learning from experts to learning from experience. This innovation can help agents extract a priori knowledge from demonstrations while reducing the detrimental effects of the poor Markovian properties of the demonstrations. Our algorithm achieves good performance in MuJoCoenvironments with limited and sub-optimal demonstrations.
\end{abstract}

\section{INTRODUCTION}

Reinforcement learning has been highly successful in many areas, meeting or exceeding human performance, in video game environments \cite{mnih2015human}, robot manipulators \cite{2018Learning}, and various open-source simulation scenarios \cite{2015Continuous}. However, reinforcement learning requires frequent interaction to collect data, which is difficult in most application scenarios. These mentioned are limited significantly by the need to explore potential policies through trial and error, which produces initial performance significantly worse than human experts. It makes RL sample inefficient and slow to converge.

One way of overcoming these limitations is to utilize human expert demonstrations to provide the agent with prior knowledge and initialize it to a higher performance level than randomly initialized networks. This approach is what we call Learning from Demonstrations (LfD) \cite{argall2009survey}, which is a subset of imitation learning that seeks to train the policy to imitate the behavior of human experts. Behavior cloning (BC) is of the LfD methods, which collect demonstrations for supervised learning and produce an agent that can generate similar eligible behavior in short training time with limited data\cite{goecks2019integrating}. BC attempts to approximate a map between state and action based on the state-action pairs from expert demonstrations and generate a policy that can mimic this behavior. Related research has shown that in areas such as uncrewed vehicles, robot control, behavior cloning on large-scale off-policy demonstration datasets can vastly improve over state of the art in terms of generalization performance \cite{codevilla2019exploring}. 

\indent Though BC techniques do allow for the relatively rapid learning of behaviors, it can suffer a domain shift between the off-line training experience and the online behavior \cite{ross2011reduction}. Especially when the demonstrations are limited and not optimal, the performance of BC dramatically decreased, even generating a negative impact in our experiments with a subset of D4RL datasets. BC faces many challenges as an effective method to improve the efficiency of RL training, such as the fact that human-generated data is not strictly Markov in nature, which often poses some challenges to the algorithm. Besides, in real-world tasks, it is often difficult to obtain optimal trajectories, so we hope that RL can get a final performance boost from sub-optimal data, but over-reliance on sub-optimal demonstrations can have bad results \cite{wu2019behavior}.  Moreover, the distribution of data may be narrow or biased such as those from hand-crafted deterministic policies, or arise in human demonstrations may cause divergence both empirically \cite{fujimoto2019off} and theoretically \cite{dulac2019challenges} \cite{fu2020d4rl}.

This work proposes the TD3 learning from generator (TD3fG) algorithm to reduce the adverse effects of demonstrations' sub-optimal and non-Markovian properties. Unlike previous approaches that use demonstrations to do supervised learning and directly initialize the agent's policy network, our method utilizes demonstrations to train a small network as a reference action generator and import the reference action into the exploration noise and loss function part of TD3. This generator can generate undesirable but informative actions in foreign states. Our method utilizes the reference action with linear decreasing weights to realize a smooth transition from imitation learning (IL) to reinforcement learning (RL). We compared this method to original TD3, BC, and some other methods that utilize demonstrations like DDPGfD. The experiment results show a significant improvement in the training speed and final performance in some MuJoCo tasks. To summarize, the main contribution of this work.

$\bullet$ Propose the TD3fG algorithm in section IV.

$\bullet$ In section VI, we took 100 sub-optimal demonstrations from the d4rl dataset and ran a series of comparisons experiments with BC and DDPGfD. 

$\bullet$ In section VII, we perform ablation experiments to investigate the contributions of the individual components in our method.

\section{RELATED WORK}
\subsection{Imitation Learning}

Behavior cloning is one of the most common forms of imitation learning, which extracts the states from demonstrations as input and learns policy through supervised learning.  It updates the policy network to minimize the mean square error or cross-entropy between the output and expert actions. Behavior cloning has achieved great success in driving, locomotion, and navigation but still struggles outside the manifold of demonstration data\cite{8463162}.

Data Set Aggregation (DAgger) interleaves between expert and learned policy to address the problem of accumulating error \cite{ross2011reduction}. Furthermore, Safe DAgger introduces a safety policy that predicts the error made by a primary policy trained initially with the supervised learning approach, without querying a reference policy \cite{zhang2016query}. The safe policy takes the observation and primary policy as inputs and returns a binary label that indicates whether the primary policy deviates from the reference policy. However, DAgger requires access to experts to fix the errors accumulated during training, which makes it less feasible \cite{kiran2021deep}.

\subsection{Reinforcement Learning}

Researchers have widely applied the reinforcement learning model in auto-driving and robotics and highly investigated it because of its high-level autonomy. Robots can learn to play PC games, swing up a cart pole, walk through the maze and avoid obstacles with bipedal mechanism \cite{silver2016mastering}\cite{levine2016end}. A renewed interest in RL cascaded from success in bipedal robots and manipulator control \cite{gu2017deep}, especially for deep RL, which use large neural networks as function approximators to learn the robots' control law from the interaction with simulation environment.

Motion control of legged robots is a traditional algorithmic difficulty. RL algorithm can omit complex kinematics calculations and modeling and get good performance by interacting with the environment. In RL the working environments that can be represented as a Markov Decision Process (MDP) defined by tuple 
$ (S, A, p, r, \gamma)$ 
where $S$ states and $A$  actions. $p(s_0|s, a)$ is the probability of state transfer. $r:S \times A \to R$ is a reward function and $\gamma \epsilon [0, 1)$ is the discount factor. MDP has Markovian properties, which means the conditional probability distribution of the future state of the process (conditional on past and present values) depends only on the present state; that is, given the present, the future does not depend on the past \cite{boyen2013tractable}.Let $(\Omega ,{\mathcal {F}},P)$ be a probability space with a filtration $(\mathcal{F}_s,\ s\in I)$, for some (totally ordered) index set $I$; and let $(S,\mathcal{S})$ be a measurable space. A $(S,\mathcal{S})$-valued stochastic process ${X=\{X_{t}:\Omega \to S\}_{t\in I}}{\displaystyle X=\{X_{t}:\Omega \to S\}_{t\in I}}$ adapted to the filtration is said to possess the Markov property if, for each $A\in {\mathcal  {S}}$ and each $s,t\in I$ with $s<t$,
\begin{equation}
\begin{aligned}
&P(X_{t}\in A\mid {\mathcal {F}}_{s})=P(X_{t}\in A\mid X_{s}).
\end{aligned}
\end{equation}

For the Actor-Critic method, the agent gets the output action based on the state through the policy network and the Q value for the given state and action through the critic network.

\subsection{Combine Reinforcement Learning and Behavior Cloning}

Previous works have integrated expert demonstrations to accelerate the training process like DDPGfD and BC+FineTune \cite{2017arXiv170708817V}\cite{codevilla2019exploring}. However, suboptimal trajectory data from human expert demonstrations may lack the characteristics of MDPs, making it difficult for agents to learn, even if imitative learning helps exploration.

Two recent approaches of combining imitation learning with reinforcement learning are Deep Q-Learning From Demonstrations (DQfD) and DDPG From Demonstrations (DDPGfD). DQfD imports a margin loss which guarantees the expert actions have higher Q-values than other actions. However, this method makes it difficult to improve the demonstrator policy. In DDPGfD, the demonstration is imported into the replay buffer and extends the SOTA in RL by introducing new methods to incorporate demonstrations.

DQfD and DDPGfD maintain a replay buffer that stores demonstrations and experience from interaction with the environment and draws extra $N_D$ examples from $R_D$. In further work improvement work, researchers introduce a behavior cloning loss only on the demonstration samples\cite{8463162}. This way, it can omit the pre-train step and learn from demonstration while interacting with the environment. The major limitation is sample inefficiency and the need for a large amount of experience, which is impractical except in simulation.

Our work aims to design a new imitation learning algorithm to overcome sub-optimal demonstrations' adverse effects. The proportion of these two loss functions changes dynamically. Unlike behavior cloning which directly copies the pre-trained model weights and fine-tunes the model, our method uses the pre-trained policy network as an independent agent and generates a residual input. The residual input can be seen as a sub-optimal action and used to calculate BC loss. This method significantly improves the final performance with limited sub-optimal expert demonstration.

\section{BACKGROUND}\
\subsection{Reinforcement Learning}

We use a neural network as an approximator to map the input state to an output action. It outputs an action's probability distribution or a deterministic action. The decision-making process that determines actions based on the environment is a Markov Decision Process. We aim to pick optimal actions in every state and maximize the rewards over discrete time steps in an environment. At every time step $t$, the agent observes a state $s_t$, take action $a_t$, receive a reward $r_t$ then goes to the next state $s_{t+1}$.

The agent learns a policy that maximizes the total reward, and the Q value can represent the expectation of future accumulated rewards. The critic network has state $s_t$, action $a_t$ and a Q value of $Q_t=\sum_{i=t}^T\gamma^{i-t}r_i$ where $T$ is the terminal of the $ (S, A, r)$ sequence, $r_t$ is the reward received form every step and $\gamma$ is the discount factor for future rewards. The agent's objective is to maximize the Q value.

Researchers have developed various reinforcement learning algorithms to deal with network update problems. As for the critic network, many involve constructing an estimate of the expectation of accumulated reward for a given state $s_t$ and action $a_t$.
\begin{equation}
\begin{aligned}
    Q_\pi(s_t,a_t) &= \mathbbm{E}_{r_t,s_t\sim E, a_i\sim \pi}[Q^\pi|s_t,a_t]\\
    &= \mathbbm{E}_{r_t,s_{t+1}\sim E, a_i\sim \pi}[r_t\\
    & +\gamma\mathbbm{E}_{a_t\sim\pi}[Q^\pi(s_{t+1}, a_{t+1}]]
\label{E1}
\end{aligned}
\end{equation}

The critic network's object is to approximate the action-value function $Q^\pi$, and Bellman Function shown above can be used to estimate the actual Q value and update the network. \\

\subsection{DDPG and TD3}

Our method is developed based on Twin Delayed Deep Deterministic policy gradient algorithm (TD3). TD3 is a deterministic, off-policy, and model-free reinforcement learning algorithm which is suitable for continuous action control \cite{fujimoto2018addressing}. It applies a twin network structure and delayed update method based on Deep Deterministic Policy Gradient (DDPG), an actor-critic that combines policy gradient update and Q value estimation. It maintains a actor function $\pi(s)$ with parameters $\theta_\pi$ and a critic function $Q(s,a)$ with network parameters $\theta_Q$. The agent learns the critic function by minimizing Bellman error, synchronously learning the policy function by maximizing the Q value.

DDPG alternately interacts with environment to collect experience and update network's parameters \cite{lillicrap2015continuous}. For every step, it selects action from $a_t=\pi(s_t|\theta_\pi)+\mathcal{N}_t$ according to current policy $\pi$ and exploration noise $N$. After executing action $a_t$ the agent observes reward $r_t$ and new state $s_t$. It then stores the transition $(s_t,a_t,r_t,s_{t+1})$ in replay buffer $R$. During the training step, the agent samples a random mini-batch $N$ transitions $(s_t,a_t,r_t,s_{t+1})$ from $R$. According to Bellman equation, the target value of critic network output is:
\begin{equation}
    y_i=r_i+\gamma Q(s_{i+1}, \pi'(s_{i+1}|\theta^{\pi'})
\end{equation}
Update critic parameters by minimizing MSE loss:
\begin{equation}
    L=\frac{1}{N}\sum_{i}(y_i-Q(s_i,a_i|\theta_Q))^2
\end{equation}
Update the policy parameters using the sampled policy gradient:
\begin{equation}
    \nabla_{\theta_Q}J \approx \frac{1}{N}\sum_{i}Q(s,a|\theta_Q)|_{s=s_i,a=a_i}\nabla_{\theta_\pi}\pi(s|\theta_\pi)|_{s=s_i}
\end{equation}
Note that the actor and critic network interact with environment during the on-policy process while the target actor and critic network is used to update network parameters. After updating on-policy networks through the above method we need to soft update the target networks:
\begin{equation}
\begin{aligned}
    &\theta_{Q'} = \tau\theta_Q  + (1-\tau \theta_{Q'})\\
    &\theta_{\pi'} = \tau\theta_\pi  + (1-\tau \theta_{\pi'})
\end{aligned} 
\end{equation}

TD3 is a more advanced method. Its main improvement is the import twin critic networks to solve the problem of Q value overestimation. It randomly initializes two critic networks and chooses the minimum of the two outputs as the output Q value.
Furthermore TD3 also adds noise to the target action. By smoothing the changes of the Q function along different actions, the policy is more difficult to be affected by the error of the Q function. After sample mini-batch $N$ transitions $(s_t,a_t,r_t,s_{t+1})$ from $R$, modify action $a$ according to
\begin{equation}
\begin{aligned}
    &\Tilde{a} \leftarrow \pi'(s|\theta_{\pi'}) + \mathcal{N}\\
    & y_i \leftarrow r_i+\gamma min_{i=1,2} Q_{i}(s_{i+1}, \Tilde{a})
\end{aligned} 
\end{equation}
Delayed update means that in the TD3 algorithm, the update frequency of the policy (including the target policy network) is lower than the update frequency of the Q function. (It recommends that the Q network update frequency is twice that of the policy network).

\subsection{Behavior Cloning}

Behavior cloning (BC) directly learns from the expert demonstration. With states from demonstrations as input and corresponding actions as target output, the agent can directly update policy network with supervised learning algorithm using MSE cost function
\begin{equation}
    L_\pi=\frac{1}{N}\sum_i(a_{demo}-\pi(s_{demo}|\theta_pi)^2
\end{equation}
BC directly approximates the map from state to act without considering the reward function. It is difficult to perform imitation learning from high-dimensional state-action space, and the standard behavioral cloning algorithm is prone to cause the distributional shift. The agent does not know how to return to the expert track state, leading to error accumulation. Besides, non-Markov and multi-modal behavior also suppress BC's performance and play a negative role, especially with poor demonstration.

\section{METHOD}

Our method combines TD3 with demonstrations in two ways to accelerate training and avoid restrictions of poor demonstrations. We use demonstrations to generate action noise and imitation loss.

\subsection{Pre-trained reference action generator}

First, we use supervised learning and demonstrations to train a BC policy neural network. We assume that the demonstrations are limited and sub-optimal. In experiments, we prepare 100 expert demonstrations for every task, and these demonstrations include relatively good trajectories, sub-optimal trajectories, and even some failed trajectories. The max, min, and mean accumulated reward of demonstrations are in the experiment part. Since the amount of data is small and demonstrations only cover part of action space and state space, we only use the BC method to train the generator for 100000 steps.
\begin{equation}
    L_\pi=\frac{1}{N}\sum_i(a_{demo}-\pi(s_{demo}|\theta_pi)^2
\end{equation}
\indent The generator can generate actions that are unreliable but with reference values. Experiment shows the generator performed unstably when directly used in tasks especially when states outside demonstrations.

\subsection{Generate exploration noise}
Second, we use the pre-trained BC network to generate action noise. In this part, we introduce the two parts of the noise in our method. TD3 trains a deterministic policy in the off-policy method. Since the policy is deterministic, the agent may not try enough branch actions to find valuable experiences at the beginning. To make the TD3 policy explored efficiently, we added a noise consisting of Ornstein Uhlenbeck process noise (OU noise) and reference action noise.

Unlike the original TD3 algorithm, we use OU noise applied in DDPG instead of Gaussian noise. For this part, the OU process can generate correlated noise which improves the exploration efficiency of control tasks in the inertial system and help the agent explore better in an environment with momentum. The noise is presented as $\mathcal{N}$ in $a \leftarrow \pi'(s|\theta_{\pi'}) + \mathcal{N}$. The differential of OU noise is:
\begin{equation}
    dn^o_t = -\theta(x_t-\mu)dt+d\sigma W_t
\end{equation}
where $\mu$ is mean value, $\theta$ and $\sigma$ are bigger than 0, $W_t$ is wiener process. It means that when the state deviates, it will be pulled close to the mean. Regardless of the Wiener process, we can directly integrate:
\begin{equation}
    n^o_t = \mu + (x_0 - \mu)e^{-\theta t}
\end{equation}
The perturbation term is Wiener process. The increment in each time interval is Gaussian distribution.
\begin{equation}
    W_t-W_s\sim\mathcal{N}(0, \sigma^2(t-s))
\end{equation}
$\sigma^2$ is the parameter of the Wiener process, which represents the variance. In the OU process, it also determines the magnification of the disturbance
\indent The second part of noise is from the pre-trained BC policy network.
\begin{equation}
    n^g_t = \pi^{BC}(s_t|\theta^{\pi^BC})
\end{equation}
The BC policy networks can generate unstable but meaningful actions compared with completely random exploration actions. For example, in Walker2D tasks, the random noise causes the robot to be more likely to fall at early steps because the agent explores action space equally in all directions. We use BC net to add a bias to the noise so that the agent can\\ explore more purposefully and change the action distribution:
\begin{equation}
\begin{aligned}
    & a_t \sim \mathcal{N}(a_t, \sigma^2)\\
    & \hat{a_t}\sim \mathcal{N}(a_t + \alpha(t)\pi^{BC}(s_t),  \sigma^2)
\end{aligned}
\end{equation}
\indent At early steps this trick can help agent collect more meaningful experience and accelerate training. But these noise will import negative effect when the agent has explored stable policy, So we need to decrease its weight along training. The total noise can be represented as:
\begin{equation}
\begin{aligned}
    n_t = \alpha(t)n^o_t + \beta(t)n^{BC}_t\\
    \alpha(t) = max(1-\frac{t}{T_1},0)\\
    \beta(t) = max(1-\frac{t}{T_2},0)
\end{aligned}
\end{equation}
We use simple linear function $\alpha(t)$ and $\beta(t)$ as noise weights to balance exploration and exploitation. Generally, T2 is bigger than T1, so the noise from the generator converges to 0 faster. It is in line with the idea that the reference actions only work well in early steps.

\subsection{Generate BC loss}

Third, we propose a BC loss which is a MSE loss between output action and reference action from generator. This loss aims to provide a prior knowledge for agent exploration. 
\begin{equation}
\begin{aligned}
    & a_t^{BC} = \pi^{BC}(s_t|\theta^{BC})\\
    & L^{BC} = (a_t^{ref} - \pi(s_t|\theta^\pi))^2
\end{aligned}
\end{equation}
At early steps imitate the generator's actions can help agent shrink exploration scope. This apart is controlled by linear decreasing weight $\alpha(t)$. Meanwhile the weight of loss from critic increase along steps. This method allows the agent to gradually switch from imitation action to explore action space.
\begin{equation}
\begin{aligned}
    &L = \alpha(t)L^{BC}(a_t) - \beta(t)Q(s_t, a_t|\theta^Q)|_{a_t=\pi(s_t)}\\
    &\gamma(t) = max(1-\frac{t}{T_3},0)\\
    &\delta(t) = min(1.2-\gamma(t),1)
\end{aligned}
\end{equation}
We import an extra hyperparameter $T_3$, which means after $T_3$ steps, the agent is no longer affected by the generator. To simplify tuning we set $T_3 = T_2 = 0.5T_1$. Since the exploration noise generated using the OU process has a decreasing weight function, the range of exploration shrinks as the number of steps increases. So perhaps focusing on the RL loss part in advance will help speed up the training. We modified the weight of the RL loss part as
\begin{equation}
    \delta(t) = min(1 + \theta -\gamma(t),1)
\end{equation}
where we take the value of $\theta$ to be 2.\\
\indent We also try to use Q-filter to decide whether to apply imitation loss. In calculating the loss function, the Q value of the reference action from the generator is obtained through the critic first, then compared with the Q value of the output action of the policy network. If the reference action has a bigger Q value, we import the BC loss between policy output action and reference action. This method has no decreasing weights and associated hyperparameters, but the loss function has no smooth transition.
\begin{equation}
\begin{aligned}
    & L^{BC} = 1_{Q(s_t, \pi^{BC}(s_t)) > Q(s_t, \pi(s_t))} (a_t^{ref} - \pi(s_t|\theta^\pi))^2\\
    & L = L^{BC}+L_Q
\end{aligned}
\end{equation}

\section{EXPERIMENT SETUP}
\subsection{Environments}

We do several Mujoco simulations to evaluate our methods, including Ant-v2, Walker2d-V2, HalfCheetah-v2, and Hopper-v2. The task of Walker2d and HalfCheetah is to make a two-dimensional bipedal robot walk forward as fast as possible, but the robot's model and reward function are different \cite{erez2011infinite}. Ant-v2's task is to Make a four-legged creature walk forward as fast as possible \cite{schulman2015high}. We control biped or quadruped robots with different joints to walk as far as possible within specific steps in all experiments. The reward function considers four aspects: forwarding reward (FR), which depends on how far the robot walks, and health reward (HR) that depends on if the robot falls after this step. Besides, the agent also receives a penalty (negative reward) from contact and control costs (TC and CC).
\begin{equation}
    r(s_t,a_t) = FW + HR - CC-TC
\end{equation}
We collect demonstrations from D4RL benchmark \cite{fu2020d4rl}. To meet our assumptions that the demonstration is limited, we only sample 100 demonstrations from it. Not all the demonstrations are successful. There are also sub-optimal and poor demonstrations whose total rewards are below average. We accept these non-optimal demos to broaden observed state space.

\subsection{Network architecture}
We use 3 layer networks with 256, 512, 256 hidden units per layer for actor $\pi$, critic $Q$ and generator $\pi^{BC}$. Use RELU activation for critic $Q$ and tanh for actor $\pi$ and generator $\pi^{BC}$. The generated exploration noise acts on forward propagation, and the network structure is shown in Figure 1.
\begin{figure}[h]
    \centering
    \includegraphics[scale=0.5]{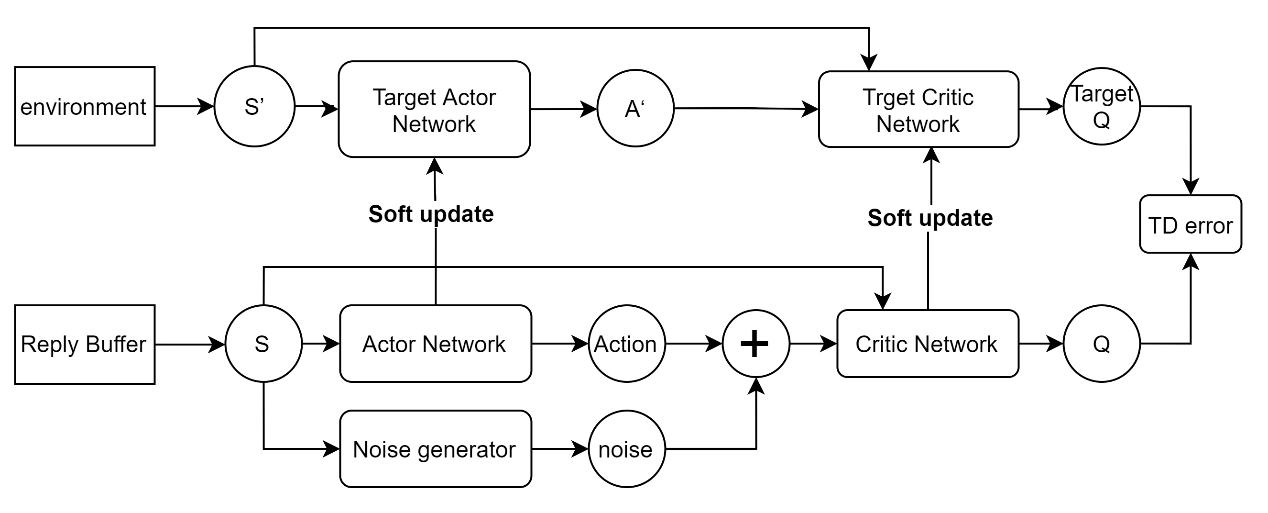}
    \caption{Caption}
    \label{fig:my_label}
\end{figure}\\
The reference action from generator effects on back propagation and its network structure is shown below.
\begin{figure}[h]
    \centering
    \includegraphics[scale=0.5]{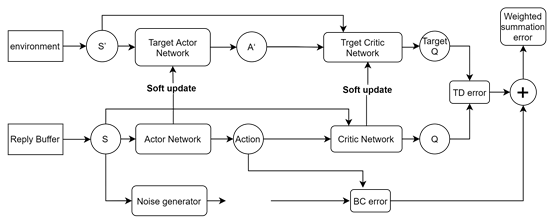}
    \caption{Caption}
    \label{fig:my_label}
\end{figure}

\subsection{Training parameters}

The discount factor $\gamma$ is 0.99. $\pi$ and $Q$ are deep neural networks with ReLU activation for $Q$, Tanh activation for $\pi$, and L2 regularization with the coefficient $10^{-4}$. We use Adam optimizer with a learning rate of $10^{-4}$. Randomly sample 640 transitions from 10 random trajectories from demonstrations and take 50000 iterations to the pre-train generator. Take 750000 steps to train the agent.

For action noise, we use OU process with $\theta = 0.15, \sigma=0.2$ to generate first part of noise. This part is weighted by a decay function $\epsilon_1 = 1 - t/6 \times 10^5$. The second part is from generator and weighted $\epsilon_2=1-t/3\times 10^5$. For loss function $T_3=3\times 10^5$.

For DDPGfD, the mean total rewards for demonstrations on replay buffer is around 5000. To ensure the consistency of the comparison experiment, we also apply the twin delay algorithm in DDPGfD.

\subsection{Overview of experiments}

We perform experiments in Walker2d, Ant, and HalfCheetah Mujoco environments and experiments with 100 and 1000 demonstrations separately. We compared our method with previous works, including original BC+fine-tune, original TD3, and DDPGfD in section VI. For comparison experiments, the training parameters are stationary. In the DDPGfD experiment, we sample 10000 transitions from the ten best trajectories and load these transitions into the replay buffer before the training begins and keep all transitions forever. 

Then we do ablation experiments to evaluate the influence of action noise, imitation loss, Q filter, and hyper-parameter $T_3$ in section VII.

\section{EXPERIMENTAL RESULT}

\subsection{Tasks}

We first show the results of our method, original BC + fine-tune and DDPGfD, on three simulated tasks:

\emph{Walker2d}: \ Make a two-dimensional bipedal robot walk forward as fast as possible. The reward function consists of forwarding reward, health reward, and control cost.

\emph{HalfCheetah}: \ Make a two-dimensional bipedal robot walk forward as fast as possible. The robot's legs are distributed on the 2D plane one after the other, unlike Walker2d. So its model is stable and does not have health rewards.

\emph{Ant}: \ Make a four-legged creature walk forward as fast as possible. Its reward function consists of forwarding reward, health reward, contact cost, and control cost.

\subsection{Baselines}

The following are three algorithms that form our baseline for comparisons.

\emph{BC and Finetunes}: Copy the weights from the generator to the agent's policy networks and interact with the environment for fine-tuning.

\emph{DDPGfD}: DDPGfD store the demonstration transitions in replay buffer and keep all the demonstration transitions during the training process\cite{2017arXiv170708817V}.

\emph{Original TD3}: Original TD3 without any demonstration information.

\subsection{Results}

The reward consists of four parts: forwarding reward (FR), health reward (HR), contact and control cost (TC and CC). The larger the total reward, the better the AGENT task is completed, moving quickly, reducing collisions, and controlling consumption while not falling over.

Figure 3 compares our work with original TD3 without demonstrations, Behavior Cloning + Fine-tune, and DDPGfD. Compared with the original TD3 model, our method significantly accelerates learning speed and obtains a better policy. It is hard for TD3 to learn to walk in walker2d and Ant without demonstrations directly. That may be due to the high dimension of action space and state space. Measuring the costs of the steps to get to convergence, our method exhibit 2x speeds up and 2x total rewards over the original TD3 in the Ant-v2 task. 

\begin{figure}[h]
    \centering
    \includegraphics[scale=0.15]{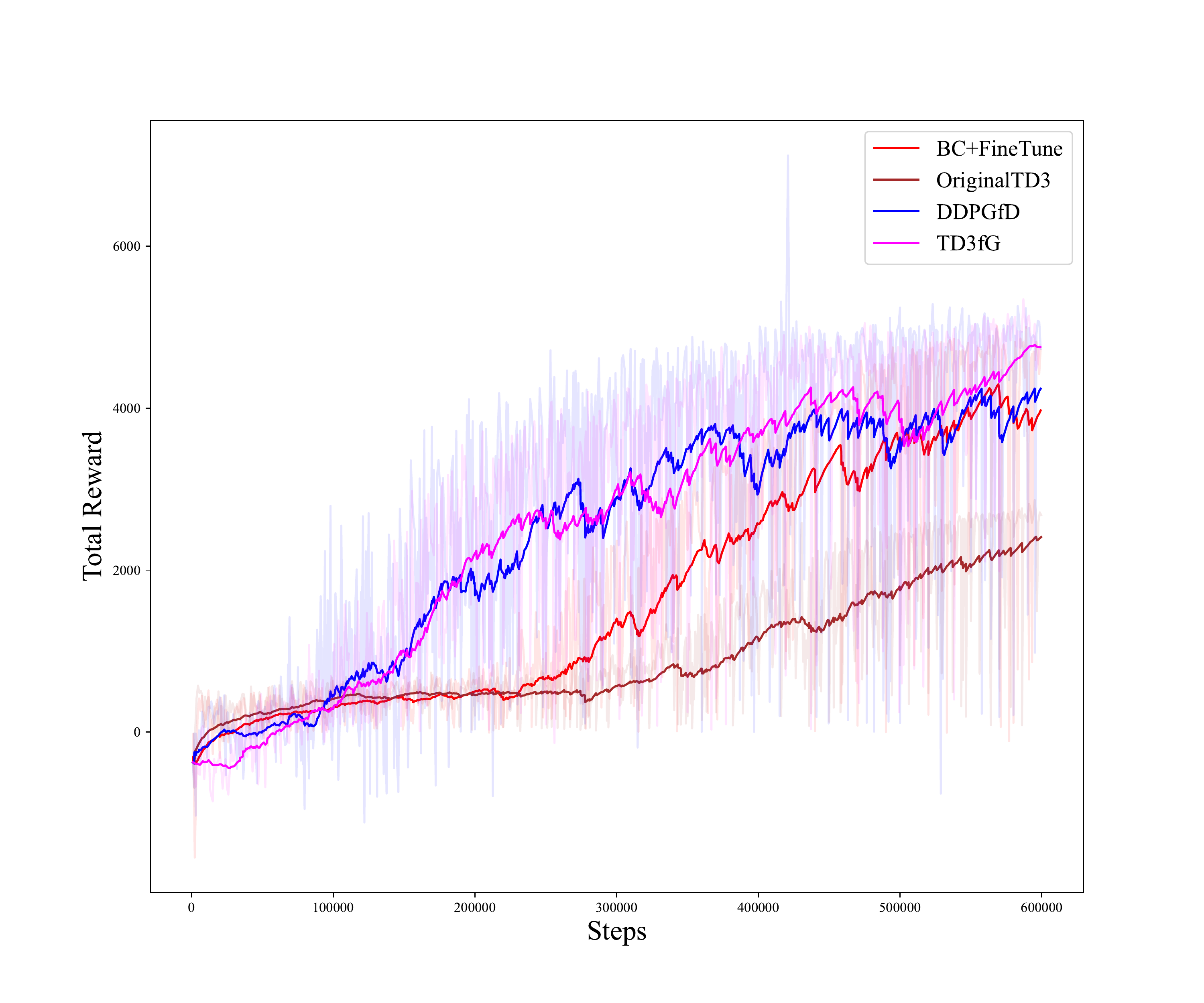}
    \includegraphics[scale=0.15]{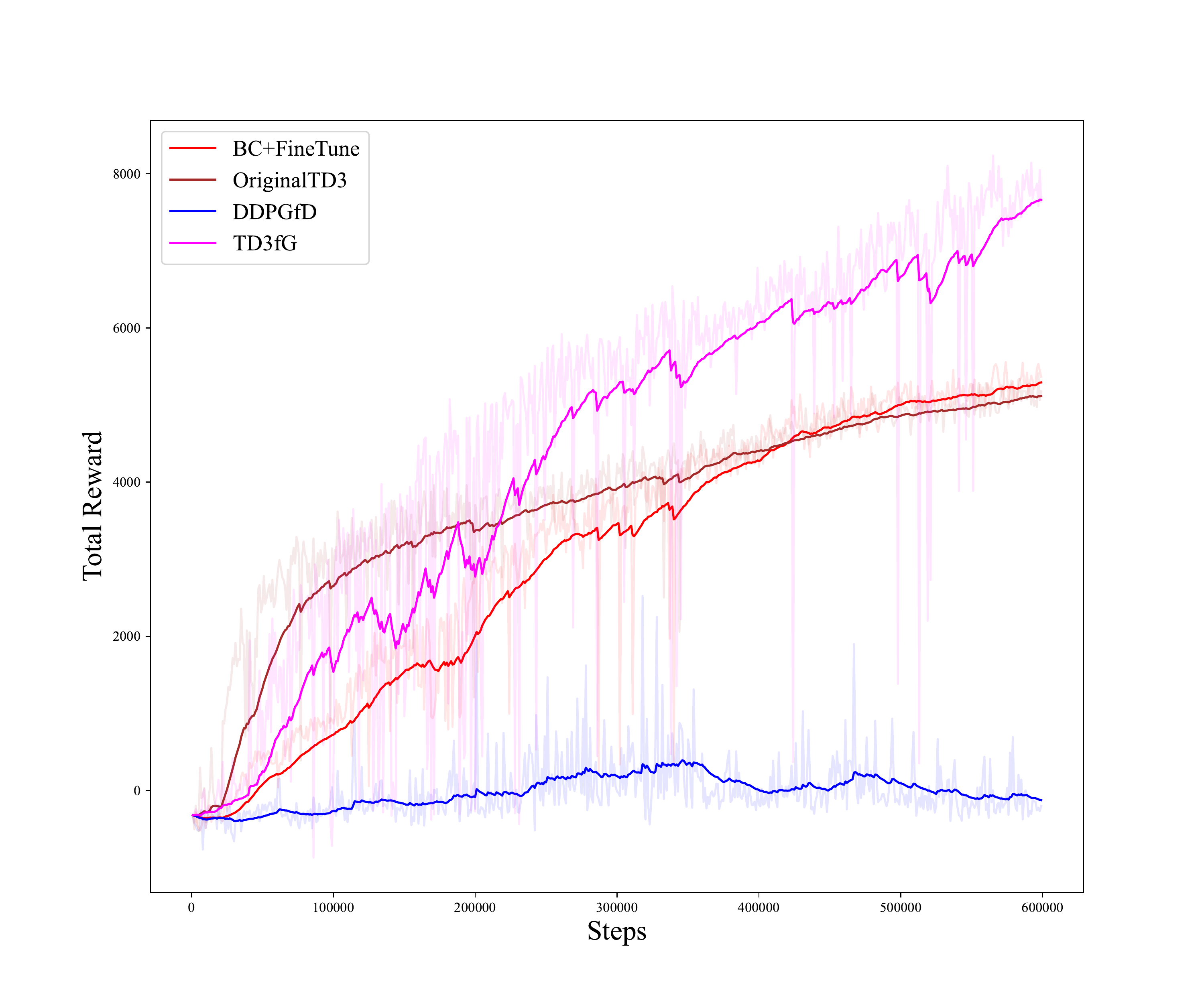}
    \includegraphics[scale=0.15]{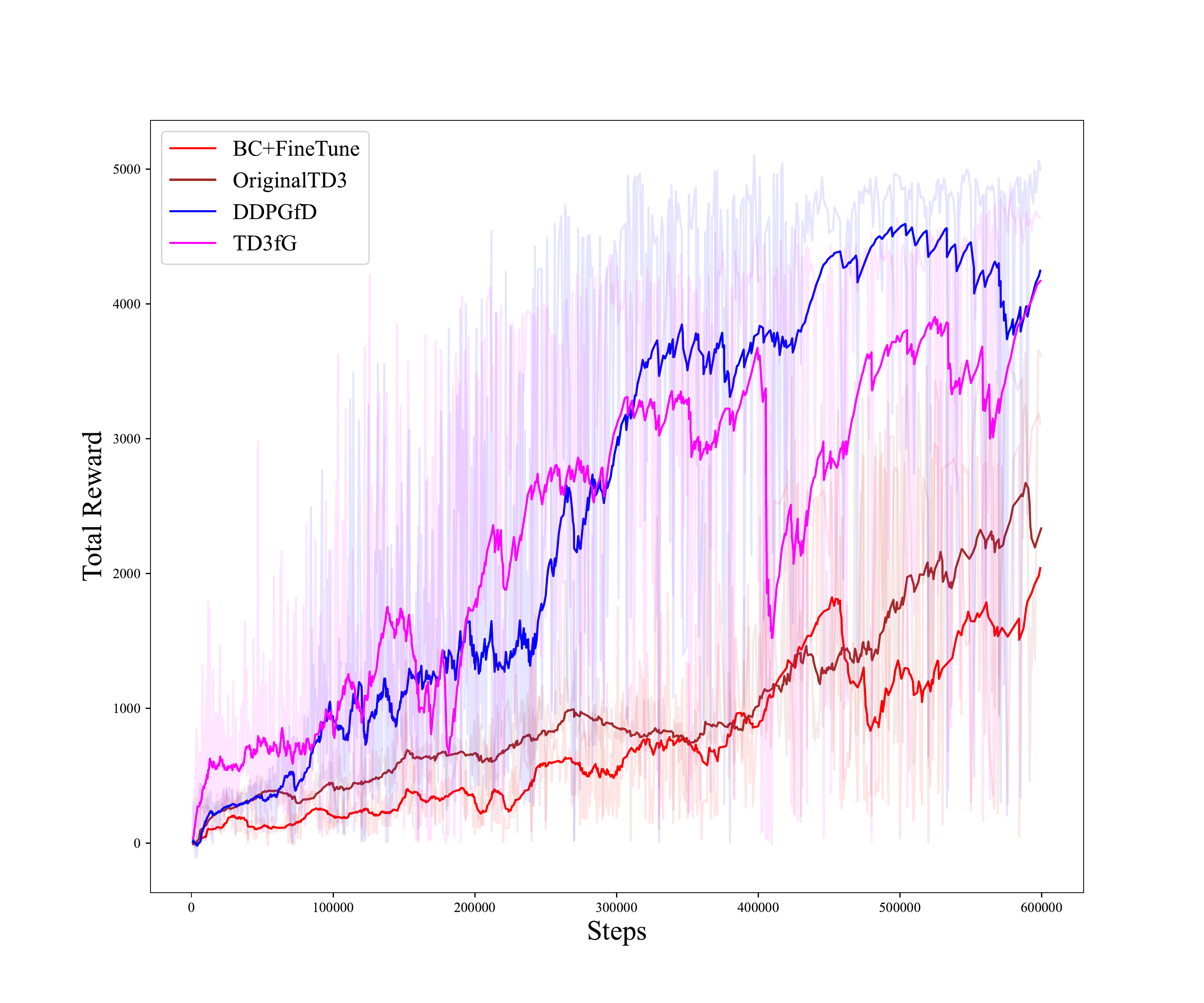}
    \caption{Comparison of TD3fG, DDPGfD and BC+Finetune}
    \label{fig:my_label}
\end{figure}

Compared with DDPGfD, our method has much better performance in HalfCheetah-v2 but not as well as DDPGfD in Walker2d-v2. In the Ant-v2 task, our approach and DDPGfD worked well, and their training speed did not differ much. The total reward of TD3fG is about 10\% higher. However, DDPGfD does not require a pre-training process but an optimized demonstration, which is unnecessary for our approach. The comparative experimental results of TD3fG and DDPGfD are in Table I.

\begin{table}[h]
\caption{comparative experimental results}
\centering
\label{table_time}
\begin{tabular}{cccc}  
\toprule   
    Tasks & Ant-v2 & HalfCheetah-v2 & Walker-v2\\  
\midrule 
    Original TD3    &2375.99    &5102.00    &2319.36\\  
    Behavior Cloning  &3960.82    &5274.89    &1943.60\\    
    DDPGfD          &4194.04    &-115.85    &4213.81\\
    TD3fG           &4704.44    &7601.85    &4065.45\\
  \bottomrule  
\end{tabular}
\end{table}

We believe that DDPGfD has a higher dependency on the quality of the expert presentation, as the presentation stored in the response buffer of DDPGfD will act on the entire training process. The adaptation to DDPGfD varies considerably between tasks. From the experiments, we can see that, especially for Walker2d, using DDPGfD even degrades the final performance. It may be influenced by the quality and quantity of the demonstrations, with sub-optimal or non-generic demonstrations continuing to negatively affect the training process, even if these experiences have good reward value.

Overall, with introducing a limited number of sub-optimal demonstrations, TD3fG outperforms Original TD3 and the traditional Behavior Cloning. At the same time, it requires less quality of demonstrations than DDPGfD, and its performance is relatively stable across tasks.

\section{ABLATION EXPERIMENT}

In this section, we perform ablation experiments to evaluate the influence of various components in our method.  We perform ablation experiments on Ant-v2, Walker2d-v2, and HalfCheetah-v2 and show the following ablations on the best performing models on each task:

\emph{Q-filter + BC loss}: In this method, it is up to the Q filter to decide whether to introduce BC loss or not. The MSE BC loss will be imported if the Q value of reference action from generator $Q(s_t, \pi^{BC}(s_t)$ is greater than $Q(s_t, \pi(s_t)$)

\emph{Decreasing weight + BC loss}: This method uses a linearly decreasing function to control the proportion of RL loss and ME loss. The weight of the BC loss gradually decays from 3.0 to 0 in the $T_3$ steps. Meanwhile, the weight of the RL loss increases from 0.2 to 1 at the same time.

\emph{BC loss + Action noise}: This method adds an exploration noise from the pre-trained generator, which we illustrate in section IV B.

\emph{BC loss + Replay Buffer}: In this method, we store 10000 sub-optimal demonstration transitions in the replay buffer and test if this helps to improve the performance. This technique is similar to DDPGfD, except it does not require the optimal demonstration in the replay buffer. 

\subsection{Q filter and Decreasing weight}

After replacing the decreasing weights with a Q filter, all three tasks' final performance and training speed drop. Figure 4 shows the training curve of TD3fG with decreasing weight and Q filter. In Ant-v2 and Walke2d-v2, the total rewards are reduced by about 20\%. The first is that the Q filter does not have a smooth transition. The second is that the critic is not reliable in the early stages of training and cannot correctly evaluate the reference movement, which weakens the role of the generator in guiding the training. These two points are not in line with the original design of the generator, so Q filter, which is a common technique, is not suitable for our method. This experiment confirms that a smooth transition between RL and IL is more helpful when suboptimal and limited demonstrations.
\begin{table}[h] 
\caption{Comparison of TD3fG and Q filter}
\centering
\label{table_time}
\begin{tabular}{cccc}  
\toprule   
    Tasks & Ant-v2 & HalfCheetah-v2 & Walker-v2\\  
\midrule 
    TD3fG           &4704.44    &7601.85    &4065.45\\
    Q filter        &1809.60    &1631.40    &3155.78\\
  \bottomrule  
\end{tabular}
\end{table}
\begin{figure}[h]
    \centering
    \includegraphics[scale=0.3]{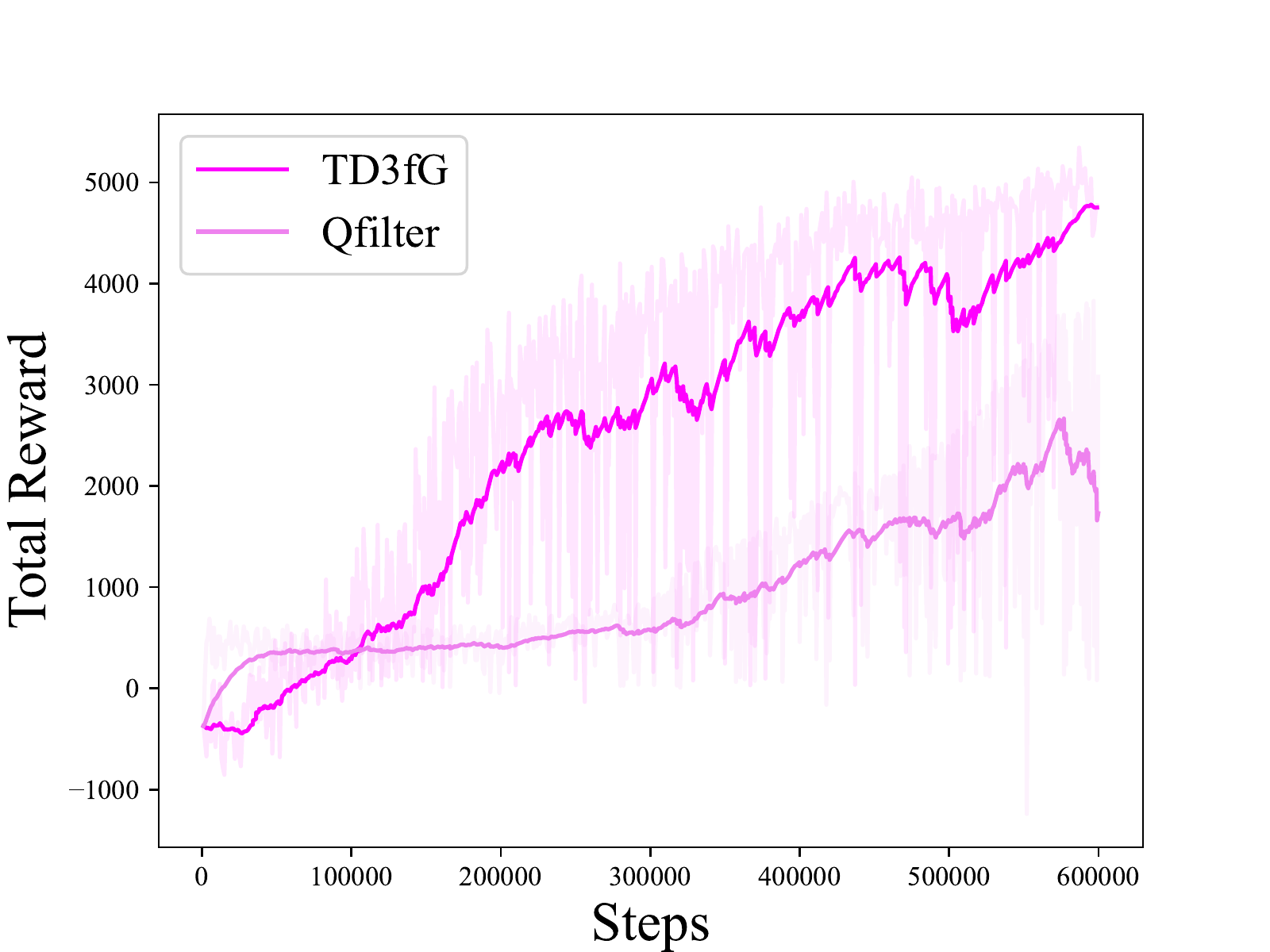}
    \includegraphics[scale=0.3]{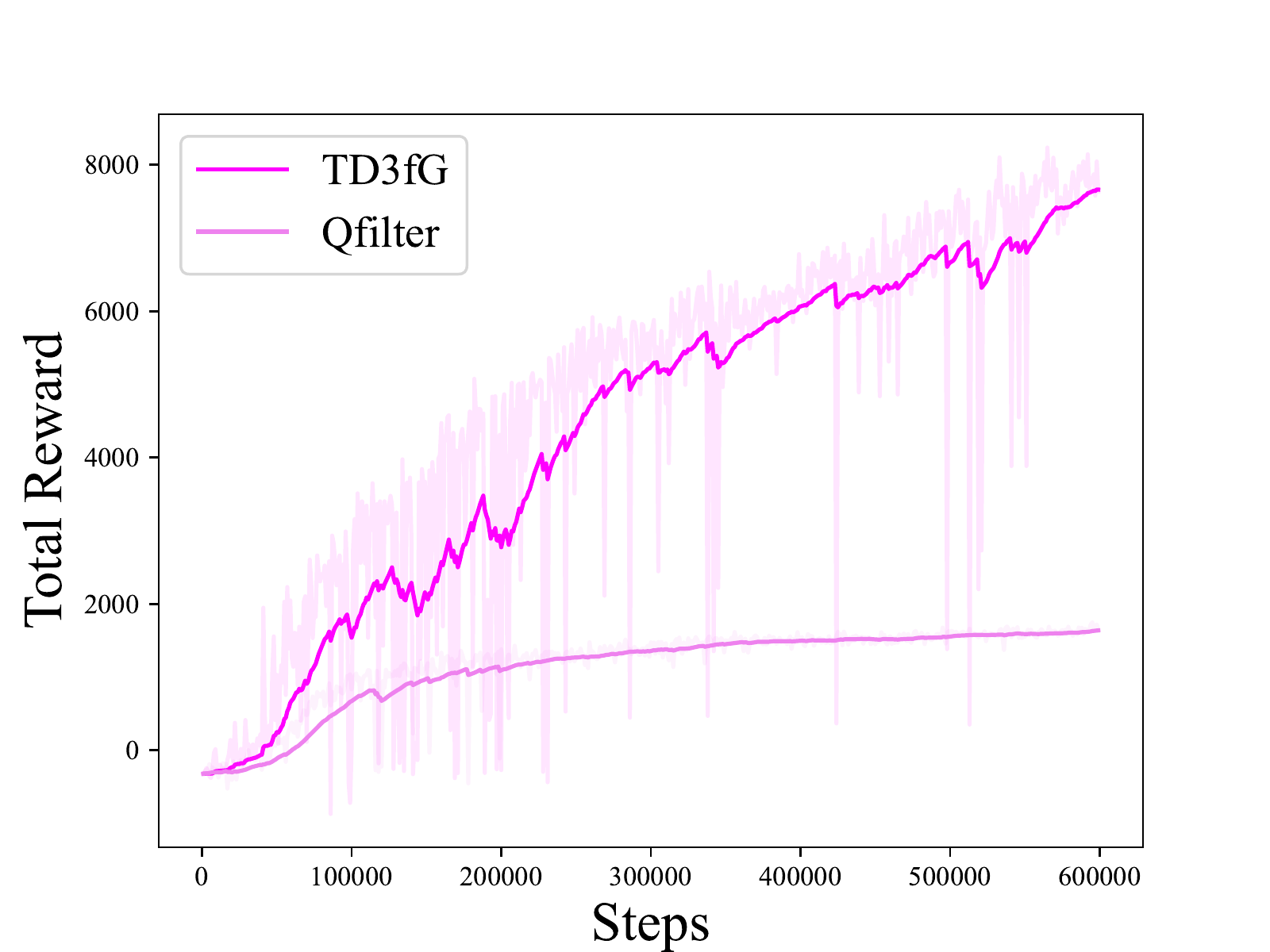}
    \includegraphics[scale=0.3]{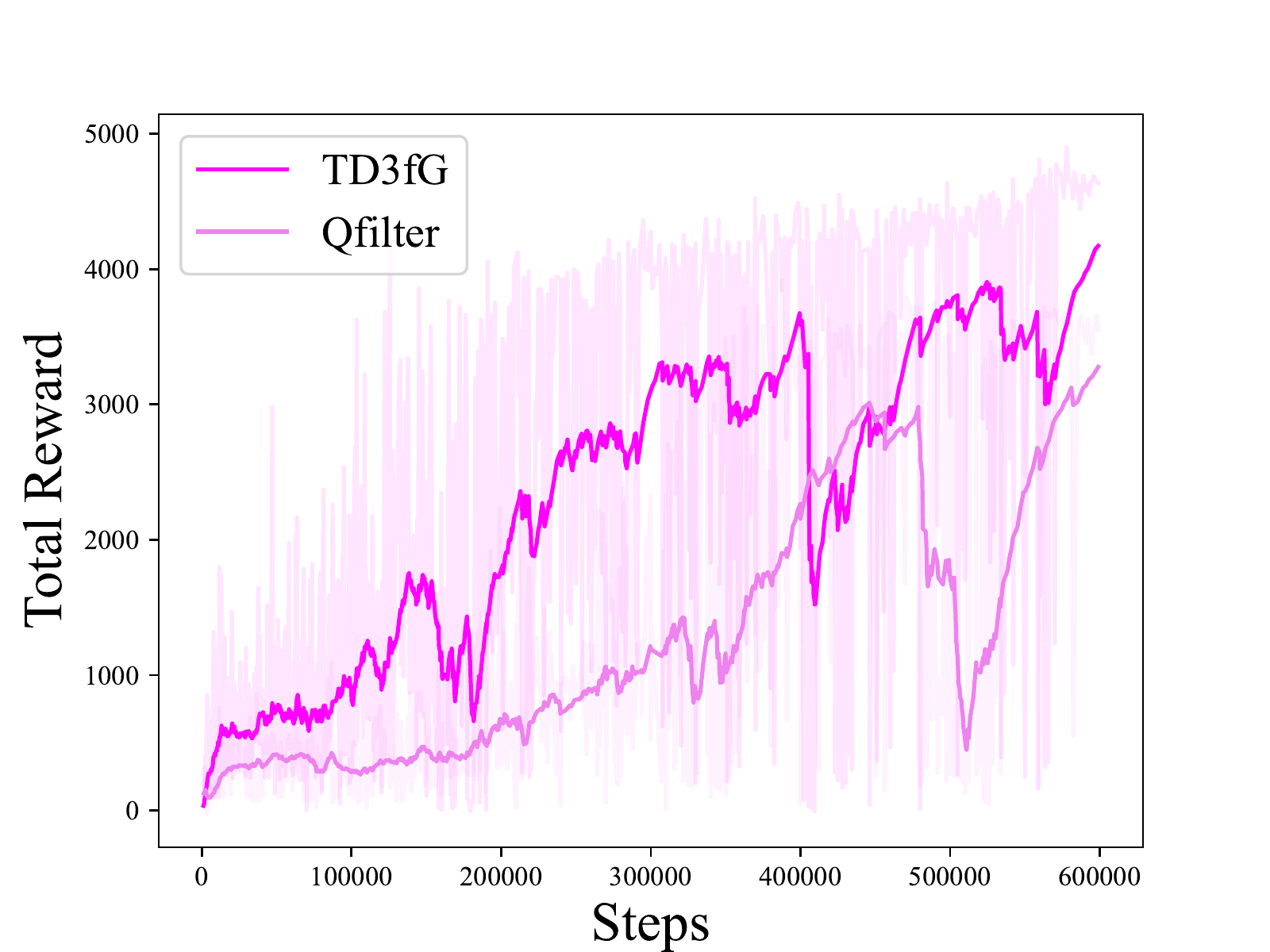}
    \caption{Comparison of TD3fG and Q filter}
    \label{fig:my_label}
\end{figure}

\subsection{Replay Buffer}

Figure 5 shows the difference between before and after using replay buffer. In Ant-v2 and Walker2d tasks, the effect of the replay buffer is negligible, and in HalfCheetah-v2, the total reward is between DDPGfD and TD3fG. It coincides with the results of previous experiments. A major difference between Td3fG and DDPGfD is the dependence on demonstrations, while TD3fG with demonstrations in replay buffer is somewhere in between in terms of sensitivity to demonstrations quality. So for the first two tasks, the performance is comparable, and in the third task, the total reward lies clearly between the two methods
\begin{table}[h]
\centering
\caption{Comparison of TD3fG and TD3fG with replay buffer}
\label{table_time}
\begin{tabular}{cccc} 
\toprule   
    Tasks & Ant-v2 & HalfCheetah-v2 & Walker-v2\\  
\midrule 
    TD3fG               &4704.44    &7601.85    &4065.45\\
    TD3fG+Replay Buffer   &4110.68    &3615.32    &4047.55\\
  \bottomrule  
\end{tabular}
\end{table}
\begin{figure}[h]
    \centering
    \includegraphics[scale=0.3]{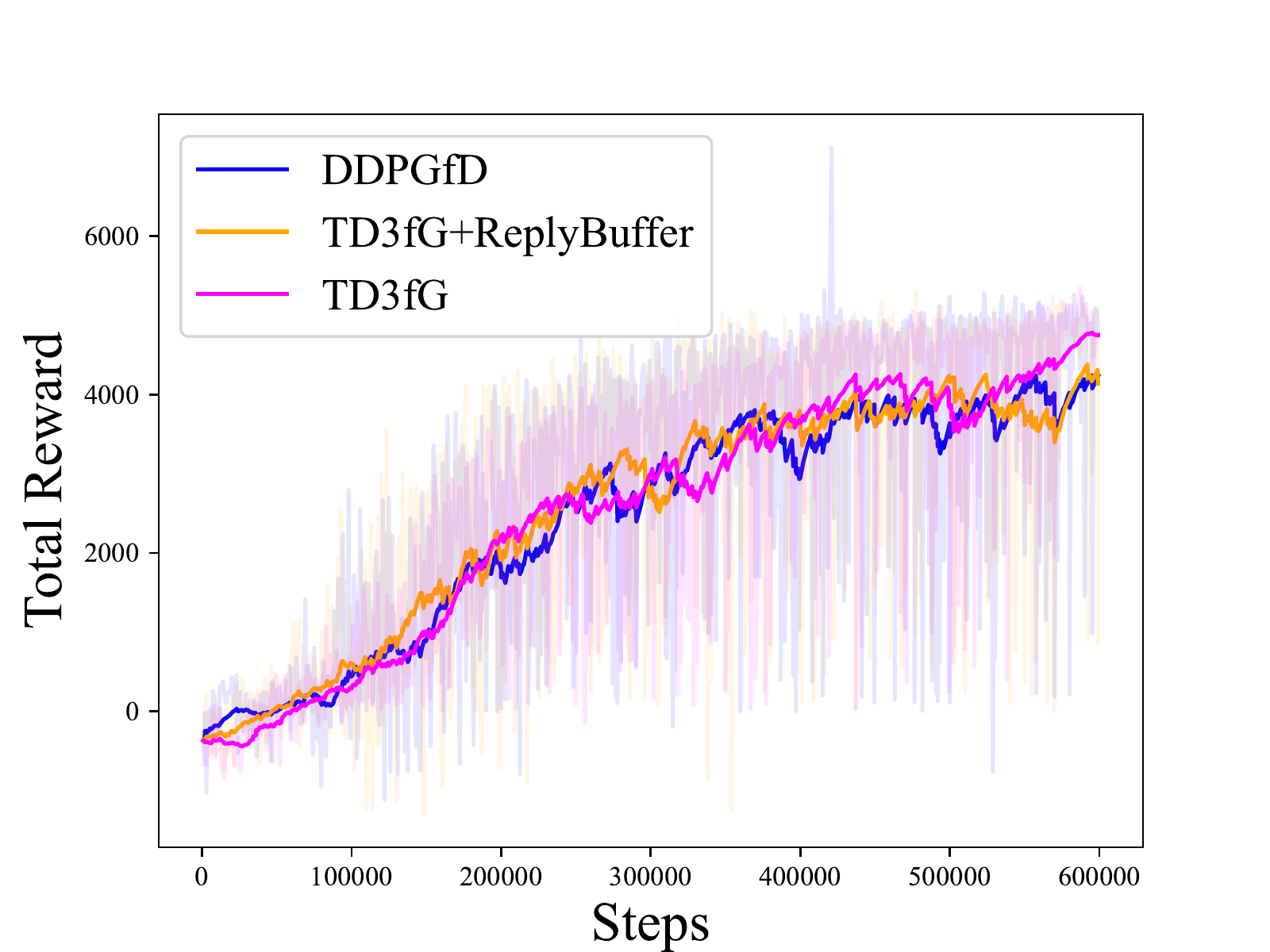}
    \includegraphics[scale=0.3]{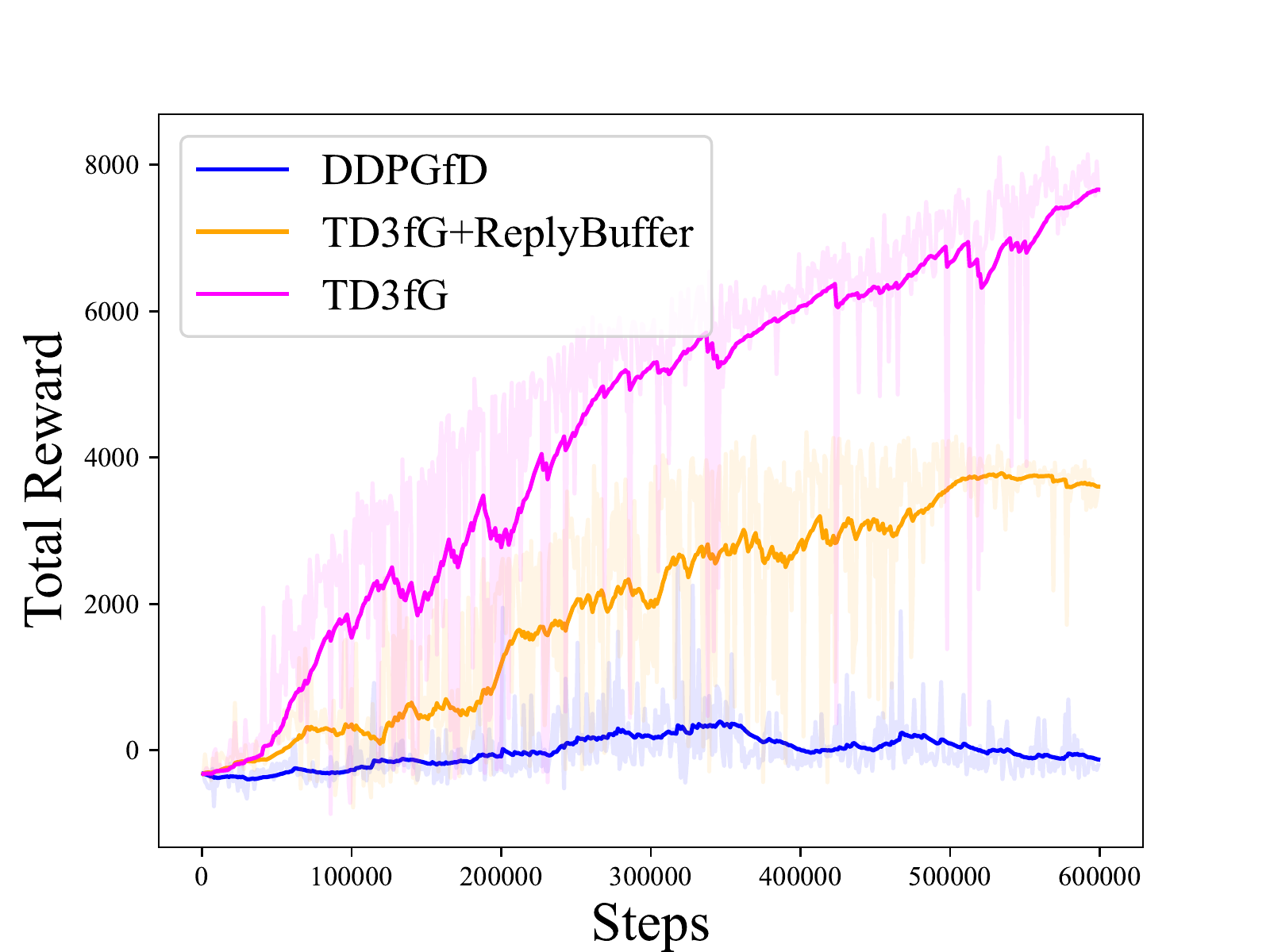}
    \includegraphics[scale=0.3]{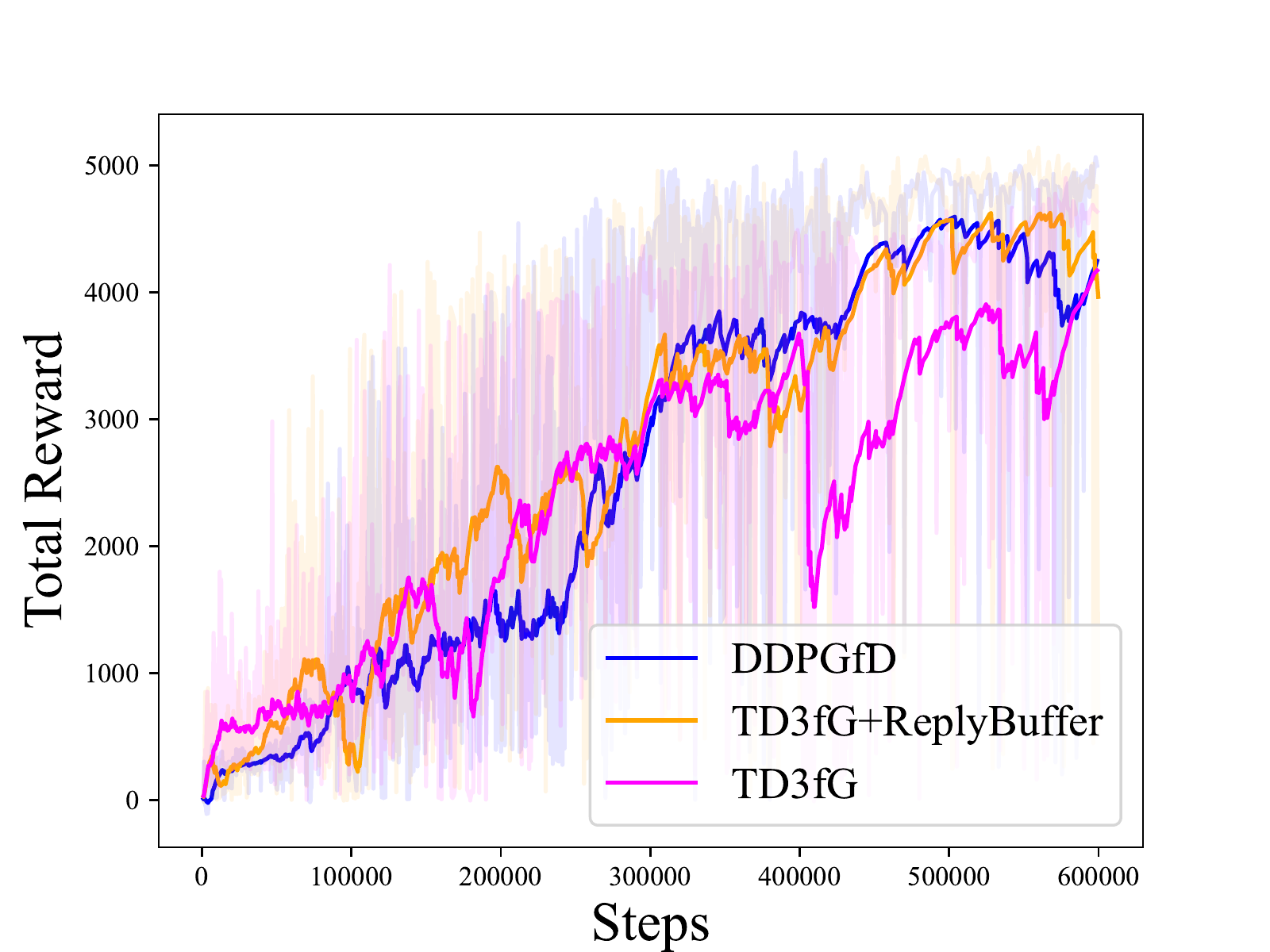}
    \caption{Comparison of TD3fG and TD3fG with replay buffer}
    \label{fig:my_label}
\end{figure}

\subsection{Action noise}

In this part we add an extra generated action noise and compare it with TD3fG that only use BC loss. For all experiments, using reference action noise alone will boost the total reward, but not as much as just using TD3fG with BC loss. Using both techniques at the same time does not give better results and may even be inferior to just using BC loss.
\begin{figure}[h]
    \centering
    \includegraphics[scale=0.3]{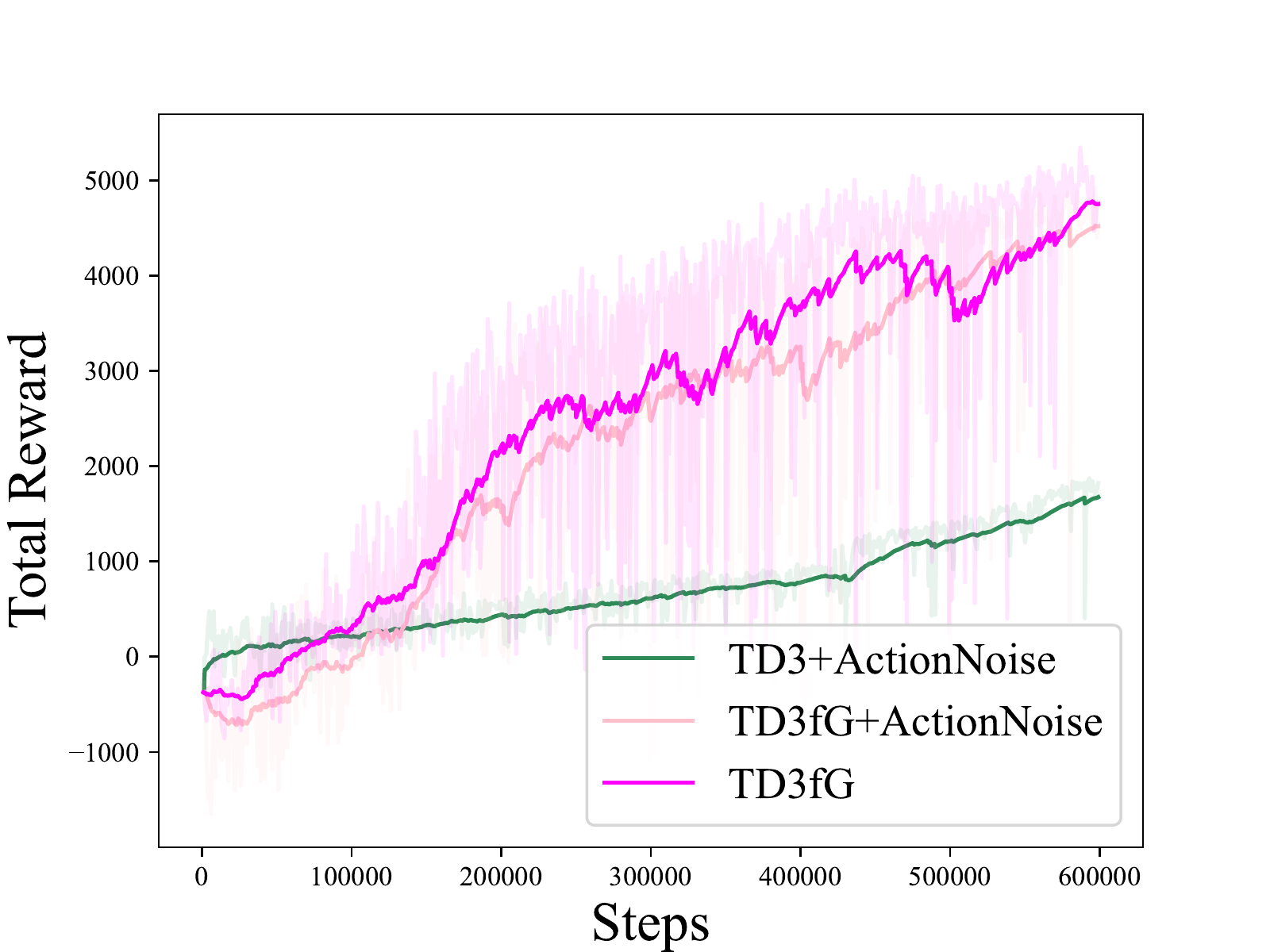}
    \includegraphics[scale=0.3]{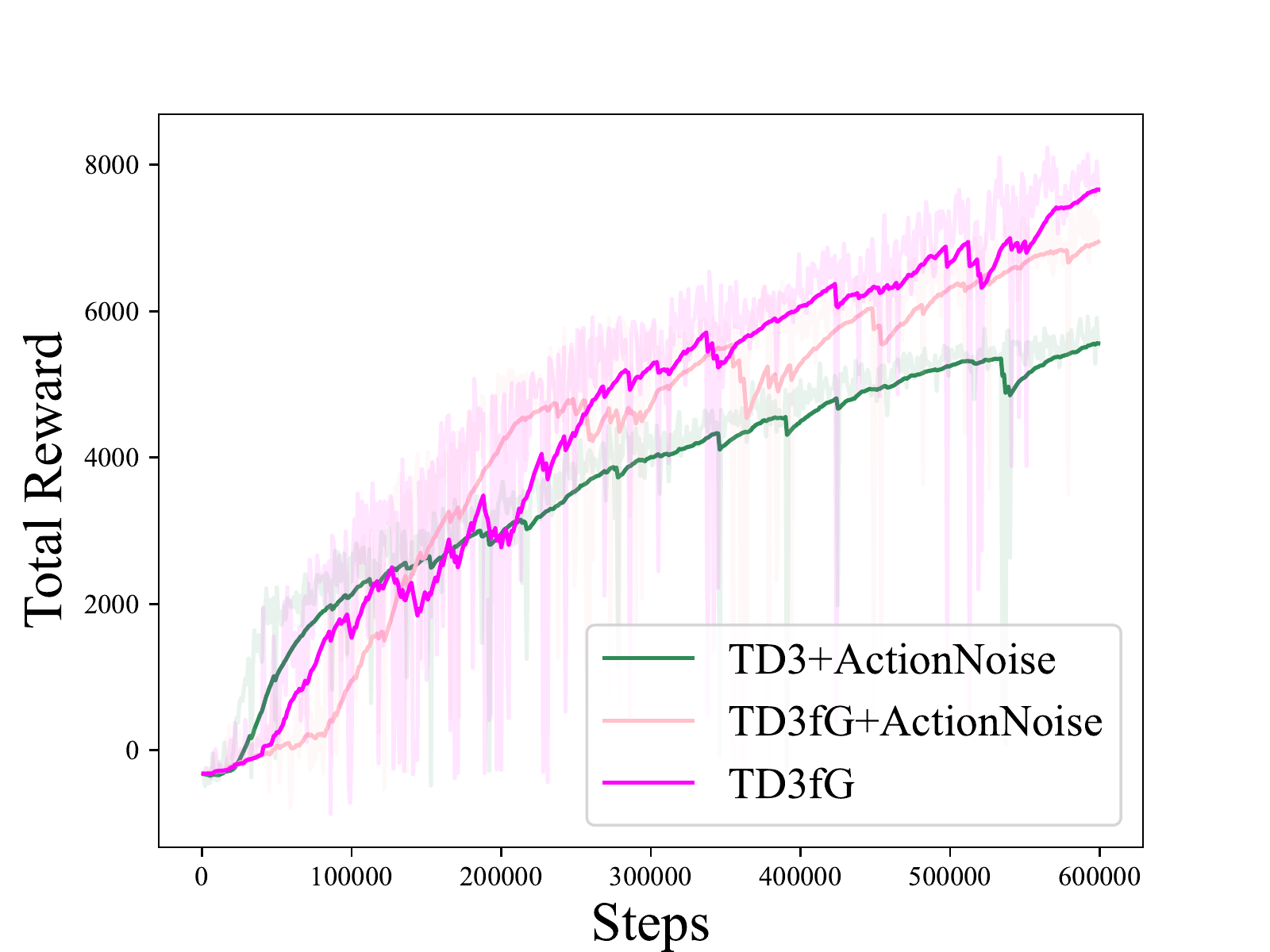}
    \includegraphics[scale=0.3]{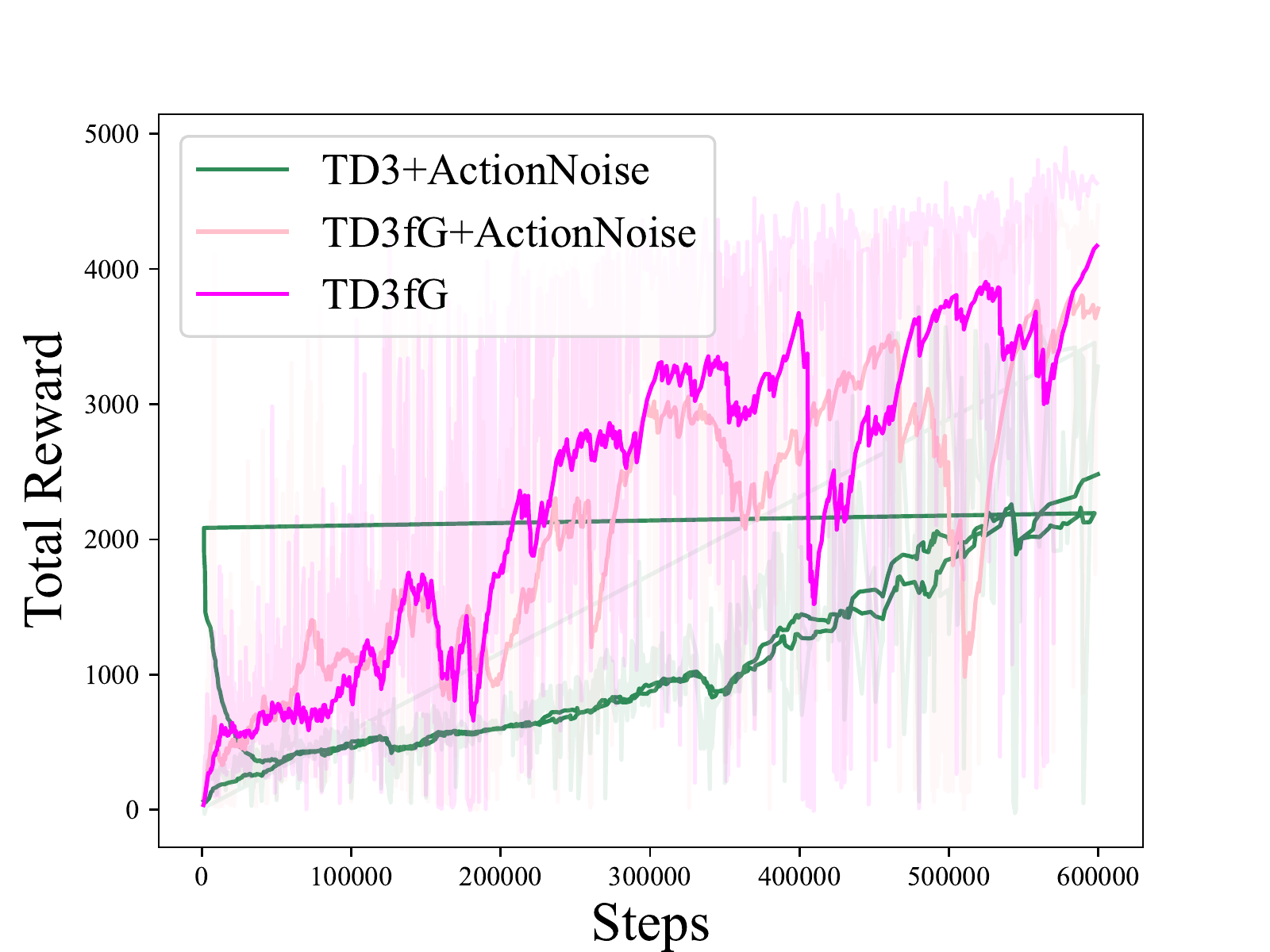}
    \caption{Comparison of TD3fG, ActionNoisea and TD3fG+ActionNoise}
    \label{fig:my_label}
\end{figure}

\begin{table}[h]
\centering
\caption{Comparison of TD3fG and TD3fG with action noise}
\label{table_time}
\begin{tabular}{cccc}  
\toprule   
    Tasks & Ant-v2 & HalfCheetah-v2 & Walker-v2\\  
\midrule 
    TD3fG               &4704.44    &7601.85    &4065.45\\
    ActionNoise   &1824.43    &5527.34    &2411.00\\
    TD3fG\\+ActionNoise &4494.95  &6913.96   &3662.40\\
\bottomrule  
\end{tabular}
\end{table}

\section{DISCUSSION AND FUTURE WORK}

We propose an algorithm that uses pre-trained generators and dynamic weights in the loss function to overcome the detrimental effects of suboptimal, non-generalized demonstrations on the reinforcement learning training process. Unlike behavioral cloning mainly for pre-training, and algorithms like DDPGfD, which use demonstrations for the entire training process, our approach achieves a smooth transition from imitation learning to reinforcement learning and performs well in simulation experiments. Our method is more general, efficient, and suitable for any continuous control task.

We conclude from the experiments that behavioral cloning is limited when the number of demonstrations is relatively tiny. The limited demonstrations are not sufficiently generalized, and there is a cumulative error. Algorithms such as DDPGfD suffer from suboptimal demonstrations when the quality of the demonstrations is low, and techniques such as Q filter allow for selective use of demonstrations. Still, they are susceptible to the critic's influence in the early training phase. Our approach uses decreasing weights with training steps, controls the ratio of RL to BC loss, is less demanding on the demonstration, and is more generalizable to different tasks.

\indent A limitation of our approach is that it takes time to train a generator in the first place, and the quality of the generator is critical to the final performance of the agent. This approach does not guarantee the agent's performance in the corner case due to the limitation of the number of demonstrations.

\bibliographystyle{ieeetr}
\bibliography{ref}
\end{document}